\documentclass[transactions,a4paper]{IEEEtran}
\usepackage{textcomp}
\usepackage{algorithm}
\usepackage{epstopdf}
\usepackage{algorithmic}
\usepackage{soul}
\usepackage{color}

\ifCLASSINFOpdf   
   \usepackage[pdftex]{graphicx}     
   \DeclareGraphicsExtensions{.pdf,.jpeg,.png,.jpg}   
 \else   
   \usepackage[dvips]{graphicx}   
   \DeclareGraphicsExtensions{.eps}   
\fi

\usepackage{fancyhdr} 

\begin{document}

\title{An Event-based Fast Movement Detection Algorithm for a Positioning Robot Using POWERLINK Communication}

\author{\IEEEauthorblockN{Juan Barrios-Avil\'es, Taras Iakymchuk, Jorge Samaniego, Alfredo Rosado-Mu\~noz}
\thanks{Juan Barrios, Taras Iakymchuk, Jorge Samaniego and Alfredo Rosado-Mu\~noz are with the Universitat de Valencia, 46100 Burjassot, Valencia. Spain. (e-mail: Juan.Barrios@uv.es).
Copyright (c) 2017 IEEE. Personal use of this material is permitted. However, permission to use this material for any other purposes must be obtained from the IEEE by sending a request to pubs-permissions@ieee.org.}
}
\maketitle
\begin{abstract}
Event-based cameras are not common in industrial applications despite they can add multiple advantages for applications with moving objects. In comparison with frame-based cameras, the amount of generated data is very low while keeping the main information in the scene. For an industrial environment with interconnected systems, data reduction becomes very important to avoid network saturation and provide faster response time. However, the use of new sensors as event-based cameras is not common since they do not usually provide connectivity to industrial Ethernet buses. This work develops a tracking system based on an event-based camera. A bioinspired filtering algorithm to reduce noise and transmitted data while keeping the main features at the scene is implemented in FPGA which also serves as a network node. POWERLINK IEEE 61158 industrial network is used to communicate the FPGA with a controller connected to a self-developed two axis servo-controlled robot. The FPGA includes the network protocol to integrate the event-based camera as any other existing network node. The inverse kinematics for the robot is included in the controller. In addition, another network node is used to control pneumatic valves blowing the ball at different speed and trajectories. To complete the system and provide a comparison, a traditional frame-based camera is also connected to the controller. The imaging data for the tracking system are obtained either from the event-based or frame-based camera. The controller acts on the servos to be moved acting as a ball follower. Results show that the robot can accurately follow the ball using fast image recognition, with the intrinsic advantages of the event-based system (size, price, power). This works shows how the development of new equipment and algorithms can be efficiently integrated in an industrial system, merging commercial industrial equipment with the new devices so that new technologies can rapidly enter into the industrial field. 
\end{abstract}



\section{Introduction}
\label{sect1}
\IEEEPARstart{T}{he} amount of data transmitted through communication networks is increasing at a higher pace than the supported bandwidth. Especially in industrial environments where real-time and low-latency systems are required, the saturation of communication networks due to the addition of advanced equipment generating and trasnmitting a high amount of data can be a problem \cite{decotignie}. These cameras produce data in form of asynchronous events, \cite{Berner_Brandli_Yang_Liu_Delbruck_2013}. Data are generated only when there is a difference in light intensity received by any of the sensors in the camera (pixels) arranged in an array. Each pixel of the camera that can sense this difference in intensity will produce an event if such difference is bigger than a predefined threshold. The generated event includes information about the address of the pixel in the sensor where the threshold was exceeded, together with a time-stamp in order to generate a unique event, not just in space but also in time. It is possible to define if the event is caused by an intensity increment or a decrement, causing a positive or negative event. This behaviour is similar to a mammal brain \cite{neuron}, which leads to use neuromorphic systems \cite{spin} for further information processing, feature extraction, scene detection \cite{tracking} and filtering \cite{filter_Gotarredona, rivas}. Proper lighting is a key factor in traditional industrial vision systems since it is difficult to maintain a constant light due to a constantly changing environment. Traditional solutions required the use of specific lighting systems suited for specific applications \cite{iluminacion,led,mix_light}. Event-based cameras minimize light effects since only pixel intensity differences are considered and no need of specific light intensity is required, independently of light conditions. 

Currently, applications working with event-based cameras have been developed for research purposes, emulating a neuromorphic system \cite{vision_robotics} and only a few are targeting the industrial sector. Some works are focused on developing and improving systems for data exchange between two or more bioinspired devices \cite{spinnaker,indivieri_comm}, but in fields as medicine or biology. Event-based systems have not yet achieved the desirable spread in industrial environments to benefit from their advantages. Current event-based systems still use a high bandwidth to transmit data, higher than a typical industrial systems could handle, making their advantages being overshadowed and making conventional frame-based machine vision systems being still the used technology in industry. Nowadays, event-based processing techniques are focused in producing more and better data for pattern recognition in neuromorphic systems \cite{Camunas, Zhao} and machine learning \cite{Jimenez} rather than data processing which could ease the task of further machine learning or other classification, prediction or recognition algorithms. Taking this into consideration, an algorithm was designed and tested for processing and filtering the data from event-based cameras, achieving high data reduction ratios and keeping the main information at the scene; for this reason, we call it \textit{'Less Data Same Information - LDSI'}. This technique is based on how biological neurons work where data consist on on-off event sequences. This algorithm is fully configurable, with the main goal of providing adjustable results of filtering and data reduction depending on the final application. A variety of the factors inherent to industrial vision systems must be considered,like event rate, noise, image size, light conditions among others. The LDSI algorithm has low computation compexity and allows low power image processing in the network node and lower data transfer bandwidth when compared to a frame-based camera. This improves the response time of the overall system too. The proposed approach is aimed to globally lower the computational burden obtaining lower energy consumption and less storage resources, respectively, which is a very important issue for decentralized industrial systems. 

Nowadays, it is clear that Ethernet has taken the lead in industrial communications, offering excellent price, performance and robustness capabilities which is proven to provide a solid framework for information exchange in all industrial levels, from management to the industrial plant. Moreover, with the advent of Industrial Internet of Things (IIoT) linked with the Industry 4.0 concept, Ethernet communication will play a key role in the deployment of efficient machines, data collection and control strategies in the next generation factories. However, different protocols use Ethernet as a common physical connection (mainly, layers 1 and 2 according to the ISO/OSI model). The widely spread TCP/IP is very well known and can be used for data management and supervision in industrial levels but not in the industrial plant where controllers, sensors and actuators must exchange information in a very short time, and even more important, in a maximum guaranteed time so that fast industrial processes can be properly run. In this scenario, multiple Ethernet-based protocol proposals exist, claiming to be the fastest, more robust and industry qualified. Some examples of the most popular protocols are Ethernet/IP \cite{ethip16a}, Profinet \cite{profinet08}, EtherCAT \cite{ethercat16} and POWERLINK \cite{POWERLINK13}. All of these protocols are based on Ethernet but they have modified the way data are processed in different layers from the ISO/OSI levels so that additional data handling techniques are added to satisfy timing requirements, data traffic in the bus, etc., when compared to the widely used TCP/IP protocol. 

This aim of this work is to show the feasibility of developing new nodes for Ethernet POWERLINK networks and combine it with already existing industrial equipment to deploy a full industrial system including commercial equipment and also non-commercial additions in both new hardware and new algorithms such as an event-based camera and a self-developed 2-axis positioning robot algorithm, respectively. Thus, the way from new ideas and research oriented works can take advantage of new technologies and open industrial systems as POWERLINK to quickly migrate to industry.

To investigate the feasibility of event-based systems built with modern industrial protocols, we have tried to solve a resource- and latency-demanding task of tracking of a fast moving ball. We developed a fast, accurate and low data transfer algorithm for object tracking by using an event-based camera. The tracking of fast moving objects is not an easy task for frame-based cameras since the movement can be faster than the frame-rate, which results in missing object positions. Typical industrial solution is the use of high speed cameras with a high frame-rate, increasing data flow and requiring significant computation resources. The use of an event-based camera is a viable alternative since it can provide accurate tracking at any speed with low data flow and computational burden.

A network node based on an FPGA was developed. It includes event-based camera data retrieval, a novel bio-inspired filtering algorithm, object tracking position detection algorithm and POWERLINK data transfer protocol to serve as a controlled node. In order to prove POWERLINK capabilities, the FPGA is integrated in a network including a managing node and two more controlled nodes: a two-axis servo controller and a PLC-based distributed I/O unit. The controller manages the communication among POWERLINK nodes but also includes the real-time computation of the inverse kinematics for a self-developed two-axis robot made with two synchronous motors, and MODBUS/TCP communication with an industrial PC connected to frame-based industrial vision system for object tracking with traditional computer vision techniques, for the comparison with the event-based camera.

A brief description of POWERLINK industrial network is included in section \ref{protocoldescription}. The used materials and their interconnections are described in section \ref{sect2}. Section \ref{sect3} details the event-based system based on FPGA, included camera communication, LDSI bio-inspired and tracking algorithms. Obtained results are provided in section \ref{sect4}, including real-time performance and finally, section \ref{sect6} provides conclusions.

\section{POWERLINK IEEE 61158 industrial protocol}
\label{protocoldescription}

Ethernet POWERLINK is a protocol for communication among multiple industrial devices, machines and equipment. It is designed with the goal of being implemented from the machine level to the process level, always involving industrial plant communications. At the machine level, a high-speed response is required, while at the process level, efficiency in the transmission of large amount of data is required. Some examples are the sending of a setpoint position in a servomotor or reporting the state of a machine or complete automation system to a central supervision desk. POWERLINK is based on the seven layers defined by the ISO/OSI model, as many other protocols. However, POWERLINK uses a different communication strategy and is based on slight modifications in the layers of the model according to the needs of speed and the amount of data to be transferred. For this reason, we have conducted a series of tests in order to characterize the advantages and disadvantages offered by these protocols. POWERLINK is an object oriented protocol based on CANopen protocol \cite{Danielis}; it uses a modified master/slave model where slaves can also share information among them. The master is referred to as Managing Node (MN) and the slave as the Controlled Node (CN). POWERLINK is suited to Machine and plant decentralized structures to provide users increased flexibility for adaptations and extensions due to its full adherence to the Ethernet standard IEEE 802.3, which yields two key features for its use in decentralized environments: cross-traffic and a free choice of network topology. The protocol application layer is defined as a carrier of all CANopen mechanisms \cite{canopen} but also allows the use of other communication profiles which are out of scope in this work.

\section{Materials and Methods}
\label{sect2}
\begin{figure}[h]
     \centering
           \includegraphics[scale=0.52]{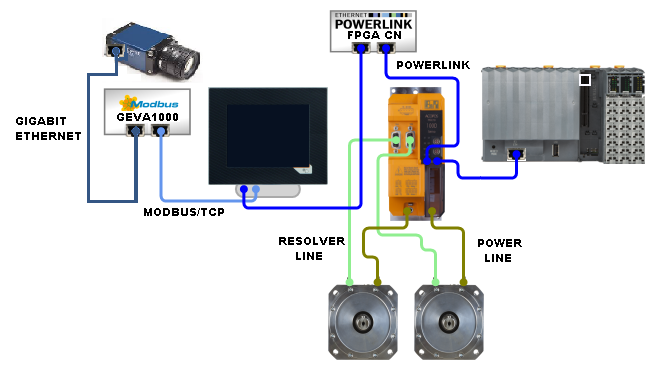}
           \caption{General view of the proposed system. Three network controlled nodes: FPGA for event-based camera, PLC for distributed I/O and 2-axis servo controller, one managing node (Power Panel controller) and one MODBUS/TCP node (PC) for the frame-based camera.}           
           \label{scheme}
 \end{figure}

The proposed system is aimed to test and compare the capabilities of a POWERLINK network to provide enough speed when the Managing Node (MN) is receiving data from an event-based camera with a self-built POWERLINK controlled node and trasmitting data to a 2-axis motor controller and a distributed I/O system; additionally, the MN is exchanging data via MODBUS/TCP through a second communication link, all simultaneously. Also, our system compares two vision technologies: traditional industrial frame-based cameras versus event-based cameras. In order to achieve these goals, a complete industrial system was built, including an industrial cabinet with two emergency stops (one located in the cabinet and the second located near the positioning robot) connected to a safety relay to meet industrial safety regulations. Additionally, electrical protections, power supplies and wiring has been done to meet industrial requirements.  The block diagram describing the equipment and interconnections used, is shown in Fig. \ref{scheme}. The following materials were used:

\begin{itemize}
\item A B\&R controller with HMI, Power Panel C70, 5.7", 1 Link interface, 1 POWERLINK interface, 1 Ethernet interface 10BASE-T/100BASE-TX and 2 USB. This panel serves as the managing node in the POWERLINK network and slave in the MODBUS/TCP network. It also provides user visualization and parameter selection \cite{BR_hmi}, together with the inverse kinematics computations for the 2-axis robot movement.
\item A B\&R PLC X20CP1382 as controlled node for distributed I/O control. Includes 14 digital inputs, 4 digital outputs, 4 digital inputs/outputs, 2 analog inputs, 2 USB, 1 RS232, 1 CAN bus, 1 POWERLINK, 1 Ethernet 10/100 Base-T \cite{BR_X20}.
\item A B\&R 2-axis ACOPOS micro 100PD.022 inverter module for servo motor control, POWERLINK interface, 2x resolver, 2 motor connections, 2 digital inputs 24 VDC \cite{BR_acopos}.
\item Two B\&R Synchronous motors, self-cooling, nominal speed 3000 rpm for the robot movement \cite{BR_servo}
\item An Anybus HMS CompactCom M40 POWERLINK gateway \cite{HMS_compactcom40}
\item A Xilinx FPGA Zedboard \cite{zedboard} for event-based camera reading, LDSI algorithm execution and POWERLINK communication. The Xilinx board acts as a controlled node in the network.
\item An event-based CMOS camera with Selective Change Driven (SCD) vision sensor \cite{pardo15}.
\item An industrial PC Teledyne DALSA GEVA1000 (2.4Ghz Dual Core, GigE x2, RS232, USB, 8 digital inputs 8 digital outputs) \cite{geva1000} connected via gigabit Ethernet with a frame-based camera DALSA Genie M640 (CR-GEN3-C6400), 1/3" format CCD with a resolution of 640x480 operating at 64 frames per second at full resolution \cite{gevagenie} and connected to the managing node controller via MODBUS/TCP. 
\item A ball movement system including a lighting control (on/off) and eight direct operated 2 port solenoid valves (SMC model VX21) \cite{smcvx} connected to the distributed I/O. The lighting consists of a white led stripe around the table where the ball is moving, controlled by digital output from the dsitributed I/O node.
\end{itemize}
 
\subsection{Managing node: Controller and HMI}

The B\&R Power Panel C70 controller is responsible for receiving data from the frame-based PC via MODBUS/TCP with the generic Ethernet port, and exchange information with the POWERLINK controlled nodes through its POWERLINK interface port. As the controller includes a 5.7" touch panel, it also serves as user interface for data visualization and parameter configuration. The controller computes the inverse kinematics equations for the robot positioning and sends the target position to the 2-axis motor drive for the robot movement according to the control positioning option: frame-based or event-based.
 
\subsection{Event-based node: SCD camera, FPGA and POWERLINK protocol}

Event-based cameras do not provide an industrial communication interface since they still are in an initial stage of development. Event-based cameras do not transmit a full frame of the sensor size periodically, they only transmit information from those intensity changing pixels. This approach results in a low data transfer rate and accurate positioning of pixels in the sensor array, and, as the intensity change is differential, a strong immunity to light exists. However, the information received from the changing pixels usually contains multiple spurious active pixels due to different noise sources from ambient conditions. This can be solved with a post-processing algorithm eliminating the pixel activity out of the main scene, thus generating less data for the post-processing algorithm as an object tracking algorithm in this case.

An FPGA is used to receive data and control and Selective Change Driven (SCD) camera using its specific electronic interface \cite{pardo15,Boluda11}. In addition, the FPGA performs a novel event-based LDSI data filtering and tracking algorithms proposed in this work (see section \ref{sect3}) and also includes all the objects and data required by the POWERLINK standard so that the FPGA can act as a controlled node by communicating with the Anybus CompactCom POWERLINK device which provides the physical POWERLINK bus connection. 

\begin{figure}[h]
     \centering
           \includegraphics[scale=.6]{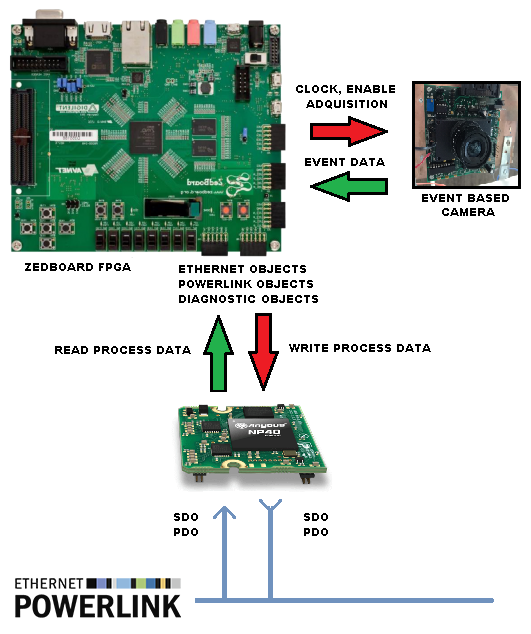}
           \caption{FPGA controlled node (CN). The FPGA reads visual event-data from the SCD camera through parallel custom signals, computes the LDSI algorithm for filtering and data reduction, performs ball tracking, and exchanges information with the POWERLINK bus through the Anybus device for physical bus connection with the MN. }
           \label{fpgaCN}
 \end{figure}

The internal connection diagram for the FPGA node is shown in Fig. \ref{fpgaCN}. The FPGA is receiving event data from the camera, the data flow and camera options is also controlled. Once event data are received by the FPGA, the LDSI filter algorithm and object tracking algorithm is performed so that the object position coordinates are obtained. Then, the FPGA is executing the communication protocol with the Anybus device so that Powerlink communication between the FPGA (acting as controlled node, CN) and the Managing Node (MN) can be done. In this case, position information is sent to the MN, which will be used to position the robot and, in turn, the MN sends to the FPGA the parameter configuration data for the filtering and tracking algorithms.

\subsection{Frame-based industrial vision system}

The GEVA1000 industrial PC receives the images from the frame-based camera, the PC includes Sherlock commercial software \cite{dalsa_sherlock} for image processing. The industrial PC and the HMI are connected by a MODBUS/TCP network where the Master node is the industrial PC and the HMI acts as slave in the MODBUS/TCP network. The 'polygonC' detection algorithm was used in this case. Algorithm configuration parameters can be adjusted from the HMI user interface (managing node) so that the user can tune the algorithm according to the specific needs. In addition to parameter configuration, the ball position can be observed in the HMI for informative purposes despite a delay is observed between the current ball position and that shown in the HMI. The delay is caused by the dynamic visualization option of the HMI despite the ball position is received with a minimum delay from the PC. The immediate availability of the ball position allows the robot positioning algorithm to calculate and position the robot with minimum delay if the user selects the control algorithm based on the frame-based vision. 

The image processing in this case performs greyscale conversion and a selection of the area of interest where the ball is moving. Then, the area of interest is pre-processed with an erode and a threshold band to remove spurious and binarize. Finally, the connectivity algorithm to find a certain blob size is applied so that the blob position (matching the ball) is obtained and converted into milimeters to inform the robot \cite{dawson14}. The HMI includes a configuration screen where the threshold and connectivity algorithm parameters can be adjusted.

\subsection{Two-axis robot}

\begin{figure}[b]
     \centering
           \includegraphics[scale=.38]{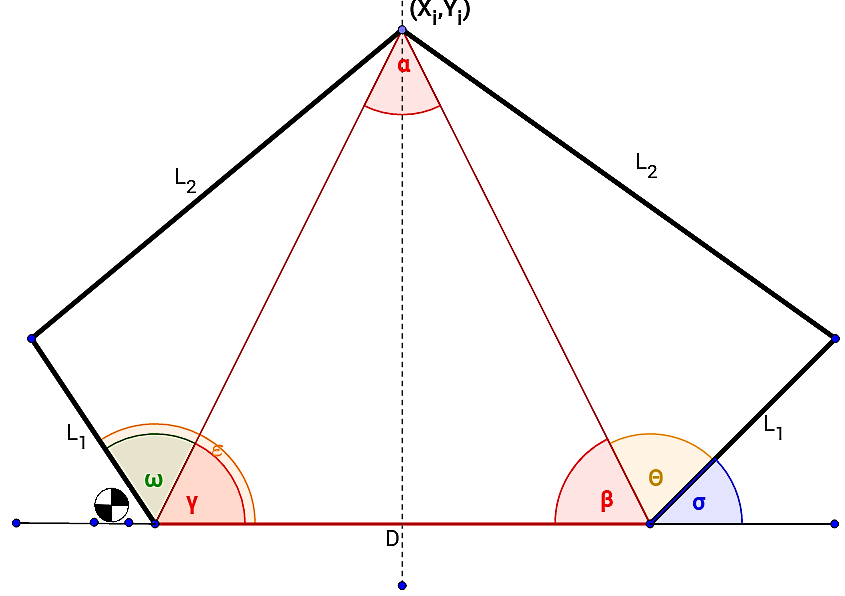}
           \caption{Two-axis robot drawing for inverse kinematics calculation.}
           \label{axisrob}
\end{figure}

The proposed robot is designed to position the Tool Center Point (TCP) in any position $(X_i,Y_i)$ of a plane in a certain range. The robot is made of two servo motors separated a distance $D$, the first motor (first axis) is located in the origin $(X,Y)=(0,0)$ and the second motor is located in $(X,Y)=(D,0)$. Each axis contains one joint, i.e. two links with length $L_1$ and $L_2$ respectively, being equal in both axis. The final link in first and second axes are connected in a joint, forming the TCP. Fig. \ref{axisrob} shows the scheme and links connection of the proposed robot system.

According to Fig. \ref{axisrob}, the position of a certain $(X_i,Y_i)$ point is separated a distance $h_1$ and $h_2$ in a straight line from the first and second motor, with an angle $\beta$ and $\gamma$, respectively. The rotation angles $\xi$ and $\sigma$ of the first and second motor respectively are computed by the eq. \ref{eq:xdef1}. The obtained solution allows to position the TCP in $(X_i,Y_i)$.
 
\begin{equation}
\begin{array}{l}
\displaystyle {h_1}^2= {X_i}^2+{Y_i}^2 \\
\displaystyle {h_2}^2= {(D-X_i)}^2+{Y_i}^2 \\
\displaystyle 
\displaystyle \gamma= \arccos\left(\frac{X_i}{h_1}\right) \\
\displaystyle 
\displaystyle \beta= \arccos\left(\frac{D-X_i}{h_2}\right) \\
\displaystyle 
\displaystyle \omega= \arccos\left(\frac{h_1^2+L_1^2-L_2^2}{2h_1L_1}\right) \\
\displaystyle 
\displaystyle \theta= \arccos\left(\frac{h_2^2+L_1^2-L_2^2}{2h_1L_1}\right) \\
\displaystyle 
\displaystyle \sigma=180-\theta-\beta \\
\displaystyle 
\displaystyle \xi=\omega+\theta
\end{array} 
\label{eq:xdef1}
\end{equation}

The robot arms were mechanically designed and created for this application using CAM-milled aluminium for rigidity. The bearings are installed in the joints to reduce friction and improve accuracy.

\subsection{Pneumatic and lighting system}

The B\&R PLC X20CP1382 serving as distributed I/O controlled node in the POWERLINK network controls the pneumatic system for air blowing to the ball. Three different ball paths are defined acoording to three respective valves sequence, which can be switched from the HMI. The digital ouputs are connected to respective relays for valve switching. One additional output is used for led lighting of the ball table.

By means of the HMI, the user can select different options for valve switching, changing speed and the sequence of activation so that different ball movements can be forced.

\section{LDSI and tracking algorithms for event-based camera in FPGA}
\label{sect3}

\begin{figure}[]
     \centering
           \includegraphics[scale=1]{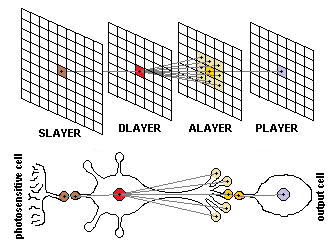}
           \caption{Top: interrelation between the different layers created in the LDSI algorithm. Lines among layer show the interconnections and data flow  from the events generated in an event sensor (SCD camera in this case) to the output layer 'Player'. Bottom: Equivalence of the proposed layer model in the algorithm with a biological neuron.}
           \label{layers_v2}
 \end{figure}

This work proposes a novel algorithm for filtering noise generated in event-based cameras and reducing the amount of redundant or irrelevant data. The algorithm, called Less Data Same Information (LDSI) has a neuromorphic basis as described in E.M. Izhikevich \cite{neuron}. The goal is not to emulate a full neuromorphic system but to take advantage of some bioinspired concepts about biological neurons in order to reduce data transmission without loss of information. The defined model and its comparison to a biological neuron is shown in Fig. \ref{layers_v2}. The layer-based neural model, similar to biological neural networks, can be observed. Units from 'Slayer' act as the dendrites feeding data to the nucleus ('Dlayer') which forward information to a wide number of synaptic terminals ('Alayer'). 

The model defines a cluster of layers serially interconnected: 

\begin{itemize}
\item The first layer corresponds to an $M\times N$ sensory layer ('Slayer') which generates the events. In this case, we have $M\times N=128\times128$ units corresponding to the number of pixels in the used SCD camera. Other event-based sensors could also be used (tactile, auditory).
\item The second layer called 'Dlayer' is a $(M-2)\times(N-2)$ units in size. This size allows to discard the border effects for the next layer and control the excitation rate of each $XY$ unit in the layer. Each $XY$ unit in 'Dlayer' is connected to its equivalent $XY$ unit and all $ij$ surrounding units in the next layer for a certain neighbourhood value or 'Depth Level, DL'. Depending on the 'DL' value, a certain $XY$ unit will be connected to the equivalent $XY$ unit in 'Alayer' ('DL'=0) or the $ij$ and surrounding units according to the 'DL' value. 
\item A third layer 'Alayer' being $(M-2)\times(N-2)$ units in size. In this layer, each $XY$ unit in the matrix receives the events generated in the $XY$ unit and all its surrounding units from 'Dlayer'. 
\item Finally, a layer called 'Player' being $(M-2)\times(N-2)$ in size provides the output. Layer 'Player' has the same structure than the input layer; this approach allows this processing system to be included in any already existing event processing system since it could interpret the output layer 'Player' as the original input layer.
\end{itemize}

All connections and layer units have multiple parameters defining their behaviour. In addition to 'DL' already explained, the following parameters related to the units in the layers are defined:
 
\begin{itemize}
\item \textbf{Excitation level of core (ERCO)}: Magnitude of the potential that a unit in the $XY$ address of 'Dlayer' increases when an event is received from the unit located in the same $XY$ address in 'Slayer'. 
\item \textbf{Excitation level of connection (ERCN)}: The potential increment in the $XY$ unit of 'Alayer' due to an event in the same $XY$ address of 'Dlayer'.
\item \textbf{Excitation level of neighbouring connections (ERNC)}: When an event is produced in an $XY$ address of 'Dlayer' this parameter corresponds to the potential increment in $XY$ unit and its neighbouring units in 'Alayer'. The number of affected units varies according to the 'DL' value.
\item \textbf{Threshold level of core excitation (TCE)}: Defines the minimum value of excitation required for an event generation in a unit in 'Dlayer'.
\item \textbf{Threshold level of connection excitation (TNE)}: Defines the minimum value of excitation required for an event generation in a unit in 'Alayer'.
\item \textbf{Maximum time to remember (MTR)}: Defines the maximum time between two events that the value of excitation in layers 'Dlayer' and 'Alayer' can remain before being degraded. This parameter can be associated to a forgetting factor in the unit.
\item \textbf{Decrement of excitation rate potential (DERP)}: The value to be decremented from the potential in 'Dlayer' once 'MTR' time has passed.
\item \textbf{Decrement of excitation rate connection (DERC)}: The value to be decremented from the potential in 'Alayer' once 'MTR' time has passed.
\end{itemize}

In addition, the behaviour can be controlled by some inherent parameters to the events as:

\begin{itemize}
\item \textbf{Actualtimestamp (AT)}: The timestamp of the actual event present in a certain connection.
\item \textbf{Lasttimestamp (LT)}: The timestamp of the previous event received in a certain connection.
\item \textbf{Deltatime (DT)}: Difference in milliseconds between the actual and the previous event coming from a certain connection. If this value is higher than 'MTR', the potential in the units is decreased.
\end{itemize}

\begin{figure}[]
     \centering
           \includegraphics[scale=.55]{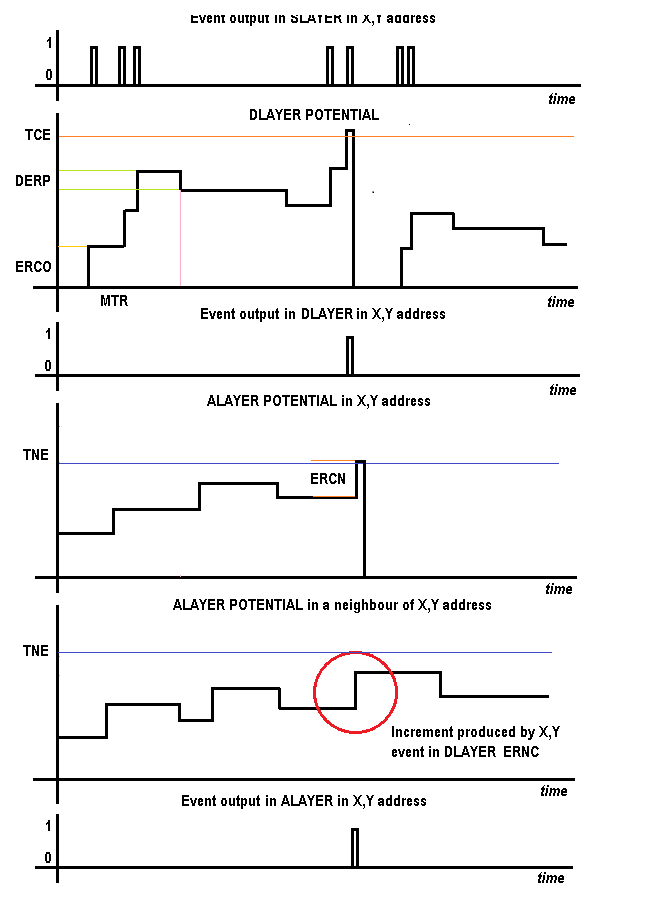}
           \caption{Event processing example for the $XY$ unit in 'Dlayer', a $XY$ unit in 'Alayer', and a neighbour unit to $XY$ upon the reception of input events. The potential value in each unit increases with input events and, when above a threshold, an output event is generated; if no events enter the unit after a time defined by 'MTR', the potential is decreased. Depending on the layer unit receiving an event from the same $XY$ unit in the previous layer, or events being received from neighbour units, the potential increment value is defined by 'ERCN' or 'ERNC', respectively.}           
           \label{par_algoritmo}
 \end{figure}

Each unit, identified by a unique address corresponding to its $XY_l$ position in the layer $l$, computes its internal potential according to the event received from its direct counterpart in the previous layer, i.e. output value from the $XY_{l-1}$ unit, and the events from the surrounding units to $XY_{l-1}$. Upon an event, the potential in the unit is increased by an excitation value which is different depending on the layer: an $XY$ unit in 'Dlayer' increases potential according to the events generated in the $XY$ unit in 'Slayer'. The potential in the unit is obtained by adding all excitation values and, when reaching a threshold, the unit generates an output event. Fig. \ref{par_algoritmo} shows an example for the behaviour of the LDSI algorithm. In this figure, input events from the sensory layer 'Slayer' in a certain $XY$ unit in 'Dlayer' are received, increasing its potential by the amount value defined by 'ERCO'. If no new events appear during an inter-event time ('DT' higher than 'MTR'), the potential is decreased by 'DERP'. When the potential at the unit reaches its threshold 'TCE', an output event is generated and transmitted to 'Alayer' which creates a potential increase by 'ERCN' at the $XY$ unit in this layer and a potential increase by 'ERNC' in its neighbours (the number of affected neighbours is given by 'DL'). When the potential value at a certain unit in 'Alayer' reaches the threshold value defined by 'TNE', an output event in the corresponding unit of 'Alayer' is generated.
 
The model defined in this algorithm makes that those events more distant in time and space to others have increased possibility of being discarded and considered as noise unless a new stimulation is received inside a certain elapsed time; depending on the connections of the neurons and their excitation levels (strengthness), an output event is generated.
 
After processing all layers, an event in a certain $XY$ unit of 'Player' is generated as long as the previous connections and core excitation levels have been strong enough to achieve a potential value above the threshold. 

\begin{figure}[]
     \centering
           \includegraphics[scale=0.24]{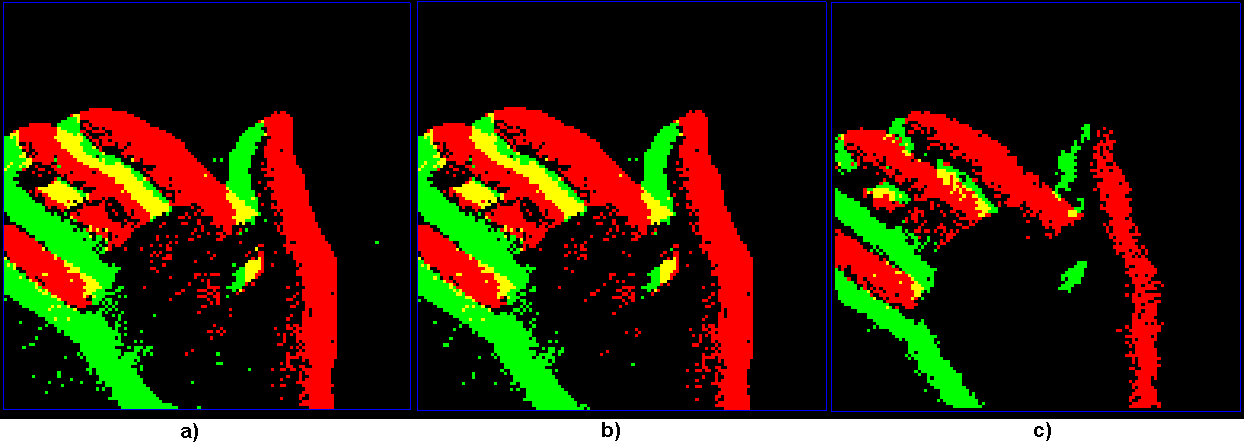}
           \caption{Low (a), medium (b) and high (c) LDSI filtering. Filter parameters allow the adjustement of output results. }
           \label{comp_filters}
 \end{figure}

Experimentally, we have determined that the range of the parameters 'ERCO', 'ERCN', 'ERNC', 'TCE', 'TNE', 'DERP' and 'DERC' should be kept between 0 and 10; otherwise, a high computational cost is required without extra benefits. The 'MTR' parameter has time units, it must be adjusted depending on the event activity (in this case, related to scene activity). Values in the order of 500 milliseconds are typical since using lower values increases the filter restrictiveness of the algorithm. Fig \ref{comp_filters} shows the results of the filter for low, medium and high filtering levels by using different parameter values.

\subsection{Tracking algorithm for the event-based camera}

After processing the event data from the SCD camera using the LDSI filtering algorithm, the relevant events corresponding to the object are obtained. As event data are continous in time, 20 consecutive events are received. These event data are computed to obtain the position for each of them and then, a vicinity algorithm is applied so that the position of the event with most neighbouring events is the winner being the latest the most valid in case of several positions having the same vicinity value.

So that this tracking provides a good performance, the LDSI filter must be fine-tuned to avoid noise influence generating false event positions.

\section{Results}
\label{sect4}

\begin{figure*}[]
     \centering
           \includegraphics[scale=0.5]{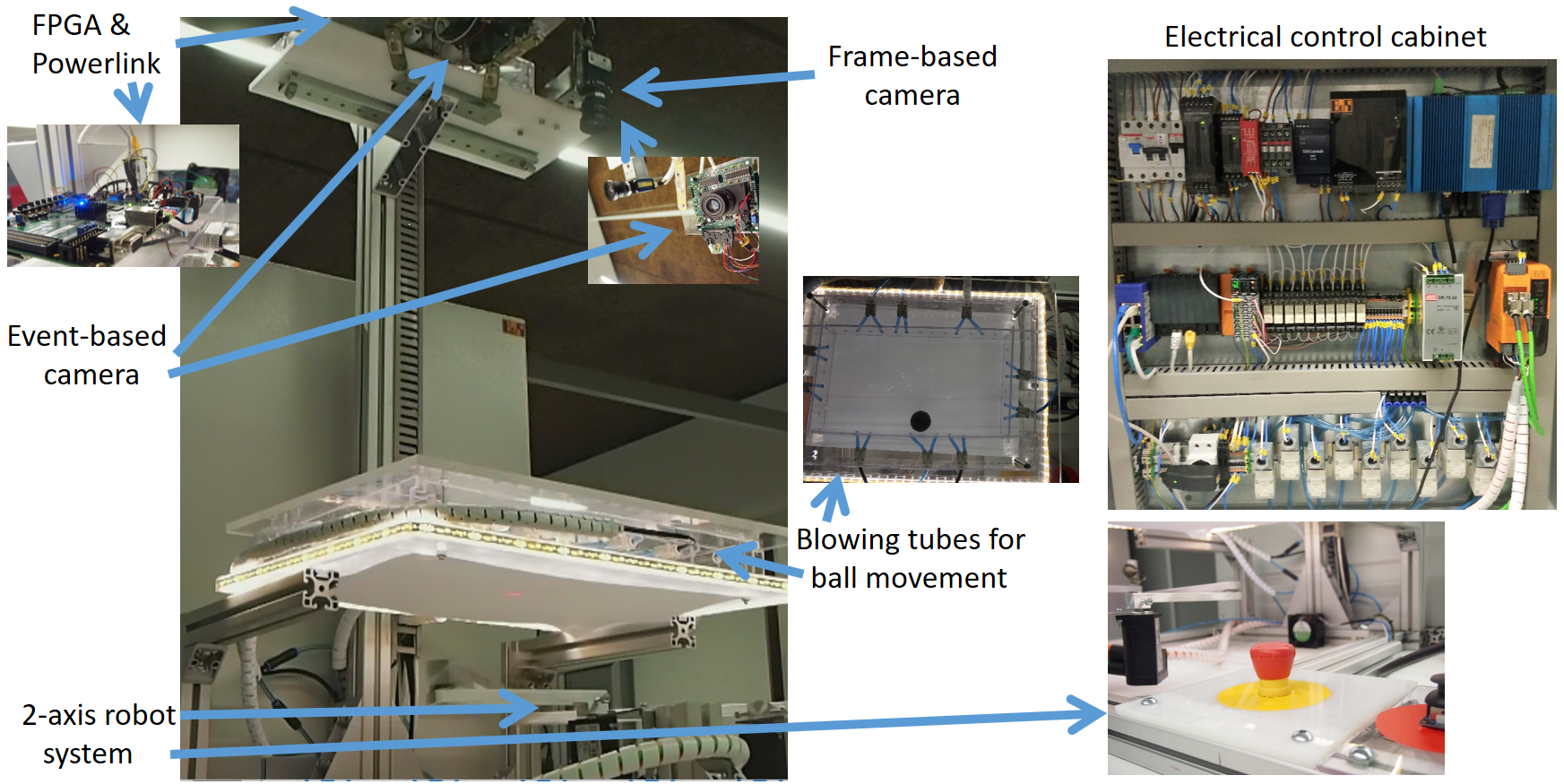}
           \caption{General view of the whole tracking system. FPGA node for event-based camera on top, frame based and event based cameras, ball movement blowing system with lighting, 2-axis robot and electrical cabinet including safety.}
           \label{comp_params}
 \end{figure*}

After the final commissioning of the system, configuration of the motor drive to accomodate the dynamics of the robot, and programming of all devices as seen in Fig. \ref{comp_params}, the testing showed that motor speed was satisfactory for fast positioning, showing stability and low vibration. 

The tracking algorithm was tested for the event-based camera. The test consisted on high, medium and low speed ball speed movement across the whole area and also in a reduced area to verify the accuracy. In all cases, the controller is able to receive the position from the FPGA node and send to the motor drive with enough speed to provide good positioning. Since no position prediction algorithm is included, there exist a delay in the actual ball position and the Tool Center Point of the robot, which is negligible in most of the cases.

In case of the frame-based camera, similar results are obtained. Thus, both technologies are able to provide good positioning, also showing that bus communications are able to provide enough speed to provide fast motor response to constant positioning changes. The main difference between evet-based and frame-based techniques lies in the fact that frame-based cameras require a high computational cost: a dedicated industrial PC with isolated Ethernet communication between PC and frame-based camera is required, image processing algorithms are computationally demanding, and power consumption is high. However, the same result is achieved by a simpler system based on a low cost FPGA with very low power consumption.

\section{Conclusion}
\label{sect6}

The integration of new devices in industry is a need in many factories where custom processes need to be carried out, and many modern tasks are requiring high-throughput communication interfaces. This work describes the sucessful integration of an event-based camera into the IEEE standard Powerlink communication bus by means of an FPGA. The camera is seen as a controlled node in the network where a commercial B\&R controller acts as managing node. In addition, the FPGA performs local computing for data filtering and object tracking, sending the tracking position to the controlled node which in turn, controls a 2-axis robot custom designed where the inverse kinematics computation for positioning is done by the managing node. Alternatively, the 2-axis robot can be controlled by a traditional frame-based camera so that a comparison between both image acquisition and tracking processing can be done. In this case, both cameras are able to provide enough speed for positioning. However, the frame based camera requires the use of a dedicated computer with isolated high bandwidth communication between the PC and the camera, resulting in a high power consumption and a more complex system while the same goal can be achieved by a custom made electronics.

\section*{Acknowledgments}
The authors would like to thank Prof. Fernando Pardo for his help and support in communitcating with the SCD camera developed by his research group at University of Valencia, and Prof. Vicente Martinez for his support in the mechanical design and manufacturing of the robot links. We would also like to thank Mr. Alejandro Cortina for his collaboration in the setting up for all the electrical system and cabinet design.


\bibliographystyle{IEEEtran}

\bibliography{IEEEabrv,references}

\begin{IEEEbiography}[{\includegraphics[width=1in,height=1.25in,clip,keepaspectratio]{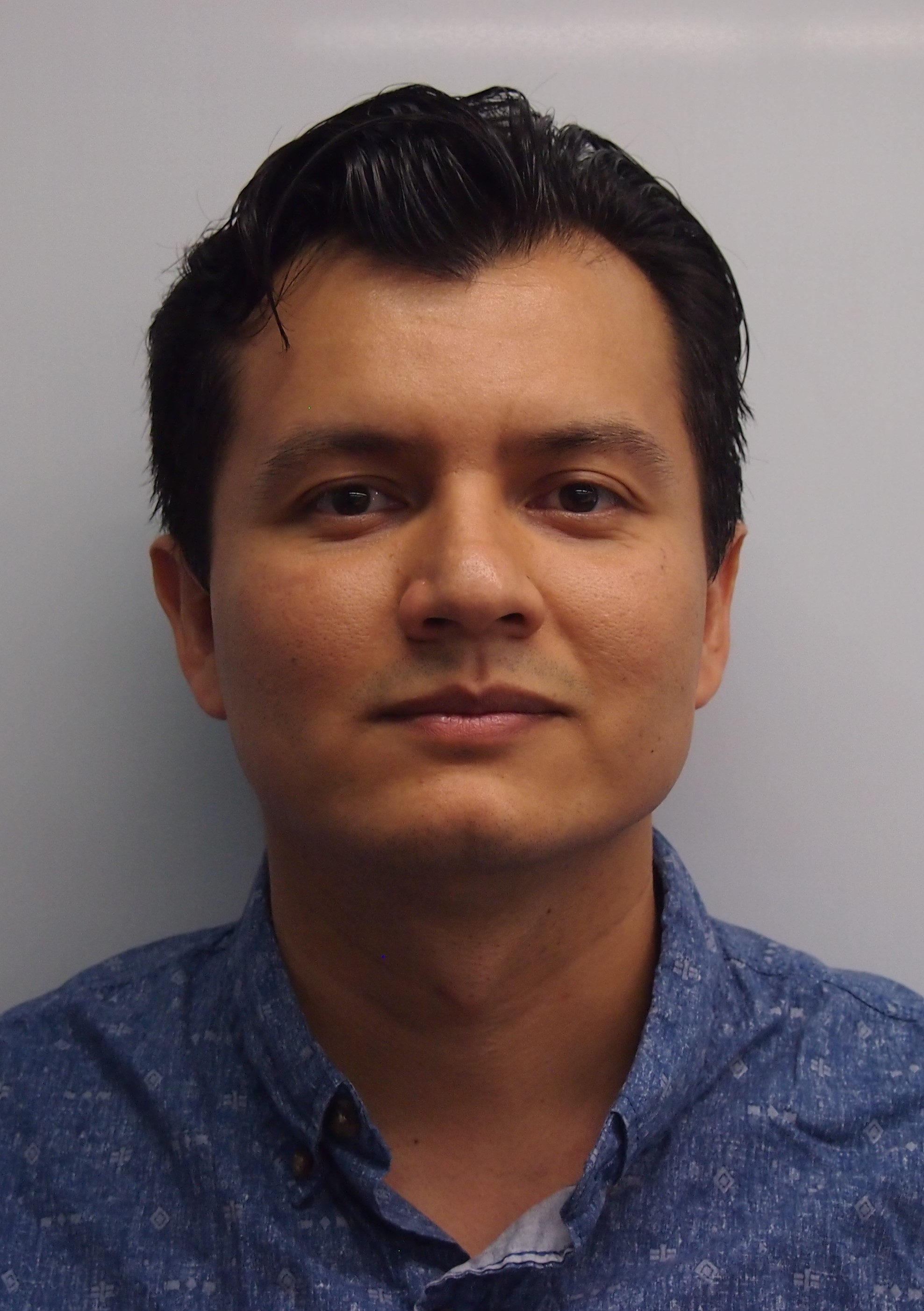}}]{\bf Juan~Barrios-Avil\'es} received a BSc degree in Mechatronic Engineering from the Instituto Tecnol\'ogico y de Estudios Superiores de Monterrey (ITESM) M\'exico and MSc in UPV. He is working towards his PhD in Electronics Engineer at the Department of electronics in the Universidad de Valencia (UV), Spain . His work involves artificial vision systems using AER methodology with application in industrial automation systems, industrial robotics design and communication and data processing generated in industrial environments.
\end{IEEEbiography}

\begin{IEEEbiography}[{\includegraphics[width=1in,height=1.25in,clip,keepaspectratio]{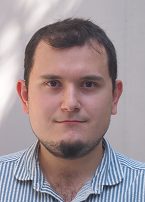}}]{\bf Taras Iakymchuk} received the MSc diploma from the Wroclaw University of Technology, Wroclaw, Poland, in 2011. He is currently working toward the PhD degree at the University of Valencia, Valencia, Spain, in the Digital Signal Processing Group. He was collaborating with research groups from Sevilla, Manchester and Institute of Neuroinformatics in Zurich. His main research interests include embedded systems, neural networks, hardware learning, and bioinspired computation.
\end{IEEEbiography}

\begin{IEEEbiography}[{\includegraphics[width=1in,height=1.25in,clip,keepaspectratio]{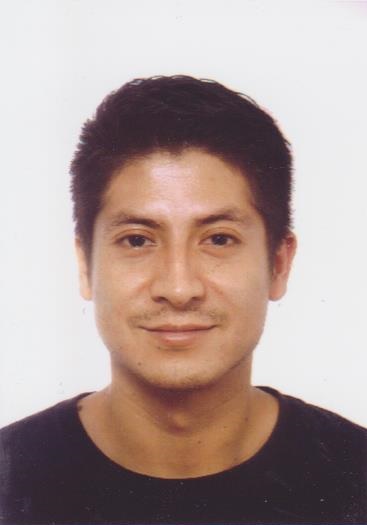}}]{\bf Jorge Samaniego} graduated at the University of Valencia in Industrial Electronics Engineering. He is a Master student in industrial automation systems. His main interests are related to PLC and HMI programming, electrical design and industrial automation in general.
\end{IEEEbiography}

\begin{IEEEbiography}[{\includegraphics[width=1in,height=1.25in,clip,keepaspectratio]{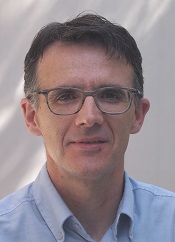}}]{\bf Alfredo~Rosado-Mu\~noz} received the MSc and PhD degrees in physics from the University of Valencia, Spain, in 1994 and 2000, respectively. He is a member of International Federation of Automatic Control (IFAC). Currently, he is professor at the Department of Electronic Engineering, University of Valencia. His work is related to automation systems, digital hardware design (embedded systems) for digital signal processing and control systems, especially targeted for biomedical engineering, and bioinspired systems.
\end{IEEEbiography}

\end{document}